\def\year{2020}\relax
\documentclass[letterpaper]{article} 
\usepackage{aaai20}  
\usepackage{times}  
\usepackage{helvet} 
\usepackage{courier}  
\usepackage[hyphens]{url}  
\usepackage{graphicx} 
\usepackage{graphicx}  
\usepackage{subcaption}
\usepackage{floatrow}
\usepackage{mathtools}
\newfloatcommand{capbtabbox}{table}[][\FBwidth]
\usepackage[breaklinks=true,bookmarks=false]{hyperref}

\usepackage{amsmath}
\usepackage[utf8]{inputenc} 
\usepackage[T1]{fontenc}    
\usepackage{hyperref}       
\usepackage{url}            
\usepackage{booktabs}       
\usepackage{amsfonts}       
\usepackage{nicefrac}       
\usepackage{microtype}      
\DeclareMathOperator{\E}{\mathbb{E}}
\usepackage{array}
\usepackage{boldline , multirow}
\newcommand{\PreserveBackslash}[1]{\let\temp=\\#1\let\\=\temp}
\newcolumntype{C}[1]{>{\PreserveBackslash\centering}p{#1}}
\newcolumntype{R}[1]{>{\PreserveBackslash\raggedleft}p{#1}}
\newcolumntype{L}[1]{>{\PreserveBackslash\raggedright}p{#1}}
\newcolumntype{M}[1]{>{\centering\arraybackslash}m{#1}}
\usepackage{xcolor}

\title{Specifying Weight Priors in Bayesian Deep Neural Networks with Empirical Bayes}

\author{%
Ranganath Krishnan\thanks{Equal Contribution} \\
Intel Labs\\
\texttt{\small ranganath.krishnan@intel.com}\\	
\And
Mahesh Subedar\footnotemark[1] \\
Intel Labs\\
\texttt{\small mahesh.subedar@intel.com}\\	
\And  
Omesh Tickoo \\
Intel Labs\\
\texttt{\small omesh.tickoo@intel.com}\\
}

\begin{document}

\maketitle

\begin{abstract}
	Stochastic variational inference for Bayesian deep neural network (DNN) requires specifying priors and approximate posterior distributions over neural network weights.  Specifying meaningful weight priors is a challenging problem, particularly for scaling variational inference to deeper architectures involving high dimensional weight space. 
	We propose \underline{MO}del \underline{P}riors with \underline{E}mpirical Bayes using \underline{D}NN (MOPED) method to choose informed weight priors in Bayesian neural networks.
	We formulate a two-stage hierarchical modeling, first find the maximum likelihood estimates of weights with DNN, and then set the weight priors using empirical Bayes approach to infer the posterior with variational inference. 
	We empirically evaluate the proposed approach on real-world tasks including image classification, video activity recognition and audio classification with varying complex neural network architectures. 
	We also evaluate our proposed approach on diabetic retinopathy diagnosis task and benchmark with the state-of-the-art Bayesian deep learning techniques. We demonstrate MOPED method enables scalable variational inference and provides reliable uncertainty quantification. 
	
\end{abstract}
\section{Introduction}
\label{sec:intro}

Uncertainty estimation in deep neural network (DNN) predictions is essential for designing reliable
and robust AI systems. Bayesian deep neural networks~\cite{neal1995bayesian,gal2016uncertainty} has allowed bridging deep learning and probabilistic Bayesian theory to quantify uncertainty by borrowing the strengths of both methodologies. Variational inference (VI)~\cite{blei2017variational} is an analytical approximation technique to infer the posterior distribution of model parameters. VI methods formulate the Bayesian inference problem as an optimization-based approach which lends itself to the stochastic gradient descent based optimization used in training DNN models. VI  with generalized formulations~\cite{graves2011practical,blundell2015weight} has renewed interest in Bayesian neural networks. 

The recent research in Bayesian Deep Learning (BDL) is focused on scaling the VI to more complex models. 
The scalability of VI in Bayesian DNNs to practical applications involving deep models and large-scale datasets is an open problem. Hybrid Bayesian DNN architectures~\cite{Subedar_2019_ICCV,krishnan2018bar} are used for complex computer vision tasks to balance complexity of the model while providing benefits of Bayesian inference. DNNs are shown to have structural benefits~\cite{bengio2013representation} which helps them in learning complex models on larger datasets. The convergence speed and performance~\cite{goodfellow2016deep} of DNN models heavily depend on the initialization of model weights and other hyper  parameters. The transfer learning approaches~\cite{shin2016deep} demonstrate the benefit of fine tuning the pretrained DNN models from adjacent domains in order to achieve faster convergence and better accuracies. 

Variational inference for Bayesian DNN involves choosing prior distributions and approximate posterior distributions over neural network weights. In a pure Bayesian approach, prior distribution is specified before any data is observed. Specifying meaningful priors in large Bayesian DNN models with high dimensional weight space is an active area of research ~\cite{wu2018deterministic,nalisnick2019dropout,sun2019functional,atanov2018deep}, as it is practically difficult to have prior belief on millions of parameters. Empirical Bayes~\cite{robbins1956empirical,casella1992illustrating} methods estimates prior distribution from the data. Based on Empirical Bayes and transfer learning approaches, we propose \underline{MO}del \underline{P}riors with \underline{E}mpirical Bayes using \underline{D}NN (MOPED) method to initialize the weight priors in Bayesian DNNs, which in our experiments have shown to achieve better training convergence for larger models.

\medskip
Our main contributions include:
\begin{itemize}
	\item  We propose MOPED method to specify informed weight priors in Bayesian neural networks using Empirical Bayes framework. MOPED advances the current state-of-the-art by enabling scalable variational inference for large models applied to real-world tasks.
	\item We demonstrate with thorough empirical experiments on multiple real-world tasks that the MOPED method helps training convergence and provides better model performance, along with reliable uncertainty estimates. 
	We also evaluate MOPED on diabetic retinopathy diagnosis task using BDL benchmarking framework~\cite{oatml2019bdlb} and demonstrate it outperforms state-of-the-art Bayesian deep learning methods. 
\end{itemize}

The rest of the document is organized as below. 
We provide background material in Section~\ref{sec_background}. The details of proposed method for initializing the weight priors in Bayesian DNN models is presented in Section~\ref{sec:MOPED}, and related work in Section~\ref{sec:related_work}. Followed by empirical experiments and results supporting the claims of proposed method in Section~\ref{sec:results}. 

\section{Background}
\label{sec_background}
\subsection{Bayesian neural networks}
Bayesian neural networks provide a probabilistic interpretation of deep learning models by placing distributions over the neural network weights~\cite{neal1995bayesian}. 
Given training dataset $D=\{x,y\}$ with inputs $x = \{x_1, . . . , x_N\}$ and their corresponding outputs $y = \{y_1, . . . , y_N\}$, in parametric Bayesian setting we would like to infer a distribution over weights $w$ as a function $y = f_w(x)$ that represents the neural network model. A prior distribution is assigned over the weights $p(w)$ that captures our prior belief as to which parameters would have likely generated the outputs before observing any data. Given the evidence data $p(y|x)$, prior distribution and model likelihood $p(y\,|\,x, w)$, the goal is to infer the posterior distribution over the weights $p(w|D)$:
\begin{equation}
p(w|D) = \frac{p(y\,|\,x,w)\,p(w)}{\int{p(y\,|\,x,w)\,p(w)}\,dw}
\label{eq:bayes}
\end{equation}

Computing the posterior distribution $p(w|D)$ is often intractable, some of the previously proposed techniques to achieve an analytically tractable inference include Markov Chain Monte Carlo (MCMC) sampling based probabilistic inference~\cite{neal2012bayesian,welling2011bayesian}, variational inference~\cite{graves2011practical,ranganath2013black,blundell2015weight}, expectation propagation~\cite{minka2001expectation} and Monte Carlo dropout approximate inference~\cite{gal2016dropout} . 

Predictive distribution is obtained through multiple stochastic forward passes on the network while sampling from the weight posteriors using Monte Carlo estimators.  Equation~\ref{eq:pred_dist} shows the predictive distribution of the output $y^*$ given new input $x^*$: 
\begin{equation}
\begin{gathered}
p(y^*|x^*,D) = \int p(y^*|x^*,w)\,p(w\,|\,D) dw \\
p(y^*|x^*,D) \approx \frac{1}{T} \sum_{i=1}^{T}p(y^*|x^*,w_i)~,~~~w_i\sim p(w\,|\,D)
\end{gathered}
\label{eq:pred_dist}
\end{equation}
where, $T$ is number of Monte Carlo samples.

\begin{table*}[t]
	\small
	\renewcommand{\arraystretch}{1.3}
	\begin{center}
		\begin{tabular}{ M{3.0cm}  M{1.2cm} M{2.3cm}  M{1.9cm}  M{1.0cm} M{1.5cm}  M{1.85cm} } 
			\hlineB{2.5}
			& & &  Bayesian DNN & \multicolumn{3}{c}{Validation Accuracy} \\ \clineB{5-7}{2}
			& & & Complexity &  & \multicolumn{2}{c}{Bayesian DNN} \\\cline{6-7} 
			Dataset & Modality & Architecture &  (\# parameters) & DNN & MFVI & MOPED\_MFVI \\\hlineB{2.5}
			UCF-101 &  Video & ResNet-101 C3D & 170,838,181 & 0.851 & {\color{red}\textbf{0.029}} & \textbf{0.867}  \\\hline
			UrbanSound8K & Audio & VGGish & 144,274,890 & 0.817 & {\color{red}\textbf{0.143}}  &\textbf{0.819} \\\hline
			Diabetic Retinopathy & Images & VGG & 21,242,689 & 0.842 & 0.843 & \textbf{0.857} \\\hline
			\multirow{ 2}{*} {CIFAR-10} & \multirow{ 2}{*}{Images} &Resnet-56 & 1,714,250 & 0.926 & 0.896 & \textbf{0.927} \\\cline{3-7}
			&  &  Resnet-20 & 546,314 & 0.911 & 0.878 &\textbf{ 0.916} \\ \hline
			MNIST & Images & LeNeT & 1,090,856 & 0.994 & 0.993 & \textbf{0.995} \\\hline
			Fashion-MNIST & Images & SCNN & 442,218 & 0.921 & 0.906 & \textbf{0.923} \\\hline
		\end{tabular}
	\end{center}
	\caption{
		\small Accuracies for architectures with different complexities and input modalities. Mean field variational inference with MOPED initialization (MOPED\_MFVI) obtains reliable uncertainty estimates from Bayesian DNNs while achieving similar or better accuracy as the deterministic DNNs. Mean field variational inference with random priors (MFVI) has convergence issues (shown in red) for complex architectures, while the proposed method achieves model convergence. DNN and MFVI accuracy numbers for diabetic retinopathy dataset are obtained from BDL-benchmarks.}
	\label{tab:Accuracy}
\end{table*}
\subsection{Variational inference}

Variational inference approximates a complex probability distribution $p(w|D)$ with a simpler distribution $q_\theta(w)$, parameterized by variational parameters $\theta$ while minimizing the Kullback-Leibler (KL) divergence. Minimizing the KL divergence is equivalent to maximizing the log evidence lower bound~(ELBO)~\cite{bishop2006pattern}, as shown in Equation~\ref{eq_ELBO}.
\begin{equation}
\begin{aligned}
\mathcal{L} := \int q_\theta(w)\,log\,p(y|x,w)\,dw - KL[q_\theta(w)||p(w)]
\end{aligned}
\label{eq_ELBO}
\end{equation}

In mean field variation inference, weights are modeled with fully factorized Gaussian distribution parameterized by variational parameters $\mu$ and $\sigma$.
\begin{equation}
q_\theta(w):=\mathcal{N}(w\,|\,\mu,\sigma)
\label{eq:posterior}
\end{equation}
The variational distribution $q_\theta(w)$ and its parameters $\mu$ and $\sigma$  are learnt while optimizing the cost function ELBO with the stochastic gradient steps.

\cite{graves2011practical} proposed fully factorized Gaussian posteriors and a differentiable loss function. \cite{blundell2015weight} proposed a Bayes by Backprop method which learns probability distribution on the weights of the neural network by minimizing loss function. \cite{wen2018flipout} proposed a Flipout method to apply pseudo-independent weight perturbations to decorrelate the gradients within mini-batches.  

\subsection{Empirical Bayes}
Empirical Bayes (EB)~\cite{casella1992illustrating} methods lie in between frequestist and Bayesian statistical approaches as it attempts to leverage strengths from both methodologies. EB methods are considered as approximation to a fully Bayesian treatment of a hierarchical Bayes model.
EB methods estimates prior distribution from the data, which is in contrast to typical Bayesian approach. The idea of Empirical Bayes is not new and the original formulation of Empirical Bayes dates back to 1950s~\cite{robbins1956empirical}, which is non-parametric EB. Since then, many parametric formulations has been proposed and used in wide variety of applications.
We use parametric Empirical Bayes approach in our proposed method for mean field variational inference in Bayesian deep neural network, where weights are modeled with fully factorized Gaussian distribution. 

Parametric EB specifies a family of prior distributions $p(w|\lambda)$ where $\lambda$ is a hyper-parameter. Analogous to Equation~\ref{eq:bayes}, posterior distribution can be obtained with EB as given by Equation~\ref{eq:EB_bayes}.
\begin{equation}
p(w|D,\lambda) = \frac{p(y\,|\,x,w)\,p(w\,|\,\lambda)}{\int{p(y\,|\,x,w)\,p(w\,|\,\lambda)}\,dw}
\label{eq:EB_bayes}
\end{equation}

\subsection{Uncertainty Quantification} 
Uncertainty estimation is essential to build reliable and robust AI systems, which is pivotal to understand system’s confidence in predictions and decision-making. Bayesian DNNs 
enable to capture different types of uncertainties: “Aleatoric” and “Epistemic”~\cite{gal2016uncertainty}. Aleatoric uncertainty captures noise inherent with observation. Epistemic uncertainty, also known as model uncertainty captures lack of knowledge in representing model parameters, specifically in the scenario of limited data. 

We evaluate the model uncertainty using Bayesian active learning by disagreement (BALD) ~\cite{houlsby2011bayesian,gal2016uncertainty}, which quantifies mutual information between parameter posterior distribution and predictive distribution. 
\begin{equation}
BALD := H(y^*|x^*, D)-\E_{p(w|D)}[H(y^*|x^*, w)]\\
\label{eq:mutual information}
\end{equation}
where, $H(y^*|x^*, D)$ is the predictive entropy as shown in Equation~\ref{eq:pred_entropy}. Predictive entropy captures a combination of input uncertainty and model uncertainty.
\begin{equation}
H(y^*|x^*, D):=-\sum_{i=0}^{K-1}p_{i\mu} * log\,p_{i\mu}\\
\label{eq:pred_entropy}
\end{equation}
and $p_{i\mu}$ is predictive mean probability of $i^{th}$ class from $T$ Monte Carlo samples, and $K$ is total number of output classes.

\begin{table*}[t]
	\small
	\renewcommand{\arraystretch}{1.3}
	\begin{center}
		\begin{tabular}{ M{2.5cm}  M{2.5cm} M{1.5cm}  M{1.85cm}  M{0.15cm} M{1.5cm} M{1.85cm} 	}	\\\hlineB{2.5}
			& Bayesian DNN & \multicolumn{2}{c}{AUPR} & &  \multicolumn{2}{c}{AUROC}   \\ \clineB{3-7}{2}
			Dataset& Archiectures & MFVI & MOPED\_MFVI & & MFVI & MOPED\_MFVI\\\hline
			UCF-101 & ResNet-101 C3D & {\color{red}\textbf{0.0174}} & \textbf{0.9186} && {\color{red}\textbf{0.6217}} & \textbf{0.9967}\\\hline
			Urban Sound 8K & VGGish & {\color{red}\textbf{0.1166}} & \textbf{0.8972}& & {\color{red}\textbf{0.551}} &  \textbf{0.9811}\\\hline
			\multirow{2}{*}{CIFAR-10} & ResNet-20  & 0.9265 & \textbf{0.9622}& &0.9877 & \textbf{0.9941}\\\cline{2-7}
			&  ResNet-56 & 0.9225 & \textbf{0.9799}& & 0.987 & \textbf{0.9970}\\\hline
						MNIST & LeNet & 0.9996 & \textbf{0.9997} & & \textbf{0.9999} & \textbf{0.9999}  \\\hline
			Fashion-MNIST & SCNN & 0.9722 & \textbf{0.9784} & & 0.9962 & \textbf{0.9969} \\\hline
		\end{tabular}
	\end{center}	
	\caption{\small Comparison of AUC of precision-recall (AUPR) and ROC (auROC) for models with varying complexities. MOPED method outperforms training with random initialization of weight priors. }
	\label{tab:AuPR}
\end{table*}
\section{MOPED: informed weight priors}
\label{sec:MOPED}
MOPED advances the current state-of-the-art in variational inference for Bayesian DNNs by providing a way for specifying meaningful prior and approximate posterior distributions over weights using Empirical Bayes framework. Empirical Bayes framework borrows strengths from both classical (frequentist) and Bayesian statistical methodologies. 

We formulate a two-stage hierarchical modeling approach, first find the maximum likelihood estimates (MLE) of weights with DNN, and then set the weight priors using Empirical Bayes approach to infer the posterior with variational inference. 

We illustrate our proposed approach on mean-field variational inference (MFVI). For MFVI in Bayesian DNNs, weights are modeled with fully factorized Gaussian distributions parameterized by variational parameters, i.e. each weight is independently sampled from the Gaussian distribution $w =\mathcal{N}(\overline{w}, \sigma)$, where $\overline{w}$ is mean and variance  $\sigma=\log(1+exp(\rho))$.  In order to ensure non-negative variance, $\sigma$ is expressed in terms of softplus function with unconstrained parameter $\rho$. We propose to set the weight priors in Bayesian neural networks based on the MLE obtained from standard DNN of equivalent architecture. We set the prior with mean equals $w_{\scalebox{.7}{$\scriptscriptstyle MLE$}}$ and unit variance respectively, and initialize the variational parameters in approximate posteriors as given in Equation~\ref{eq:mean_sel}. 

\begin{equation}
\begin{gathered}
\overline{w} := w_{\scalebox{.7}{$\scriptscriptstyle MLE$}};\,\,\,\,\,\rho \sim \mathcal{N}(\overline{\rho}, \Delta \rho) \\[2pt]
w \sim \mathcal{N}(w_{\scalebox{.7}{$\scriptscriptstyle MLE$}},\,\log(1+e^\rho))
\end{gathered}
\label{eq:mean_sel}
\end{equation}  
where, $w_{\scalebox{.7}{$\scriptscriptstyle MLE$}}$ represents maximum likelihood estimates of weights obtained from deterministic DNN model, and ($\overline{\rho}$, $\Delta \rho$) are hyper parameters (mean and variance of Gaussian perturbation for $\rho$).  

For Bayesian DNNs of complex architectures involving very high dimensional weight space (hundreds of millions of parameters), choice of $\rho$ can be sensitive as values of the weights can vary by large margin with each other. 

So, we propose to initialize the variational parameters in approximate posteriors as given in Equation~\ref{eq:scale_prior}.

\begin{equation}
\begin{gathered}
\overline{w} := w_{\scalebox{.7}{$\scriptscriptstyle MLE$}}\, ; \,\,\,\,\,\rho := \log (e^{\delta \mid w_{\scalebox{.7}{$\scriptscriptstyle MLE$}} \mid} -1)\\[2pt]
w \sim \mathcal{N}(w_{\scalebox{.7}{$\scriptscriptstyle MLE$}}, \,\,\delta \mid{w_{\scalebox{.7}{$\scriptscriptstyle MLE$}}}\mid ))
\end{gathered}
\label{eq:scale_prior}
\end{equation}
where, $\delta$ is initial perturbation factor for the weight in terms of percentage of the pretrained deterministic weight values.

\newcommand{\mysizeb}{0.19}
\begin{figure*}[h]
	\small
	\begin{subfigure}{0.34\textwidth}
		\captionsetup{
			justification=centering}
		\centering
		\includegraphics[scale=0.21]{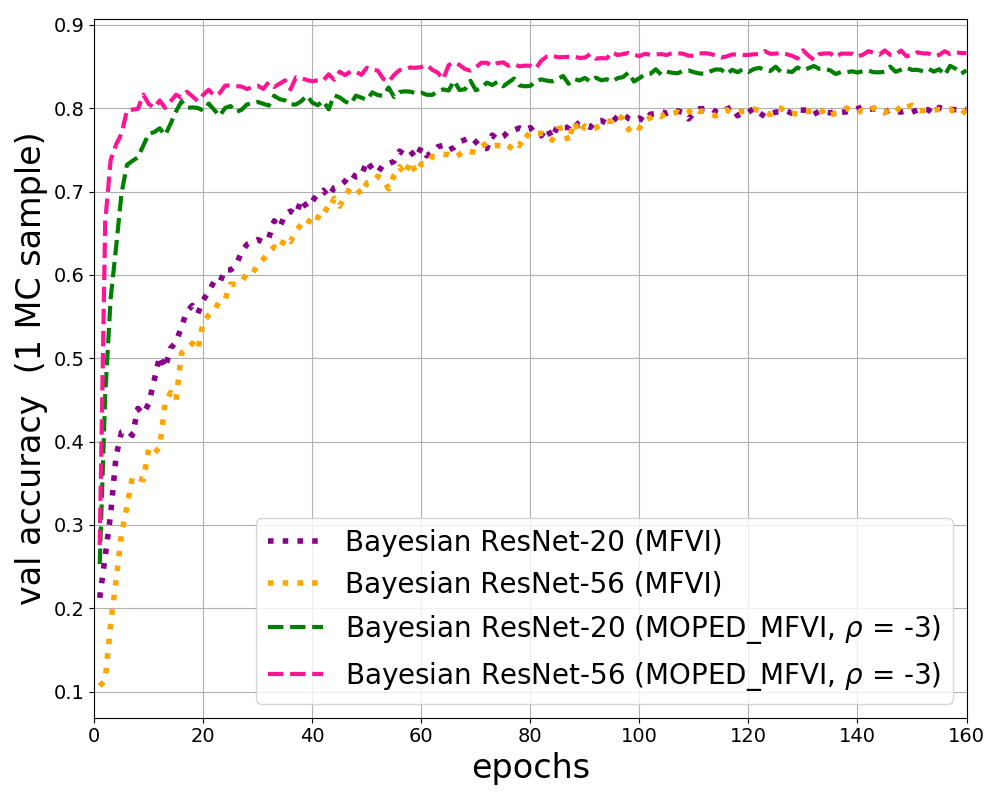}
		\caption{\small Training convergence curves}
	\end{subfigure}%
	\begin{subfigure}{0.295\textwidth}
		\centering
		\captionsetup{
			justification=centering}
		\includegraphics[scale=0.19]{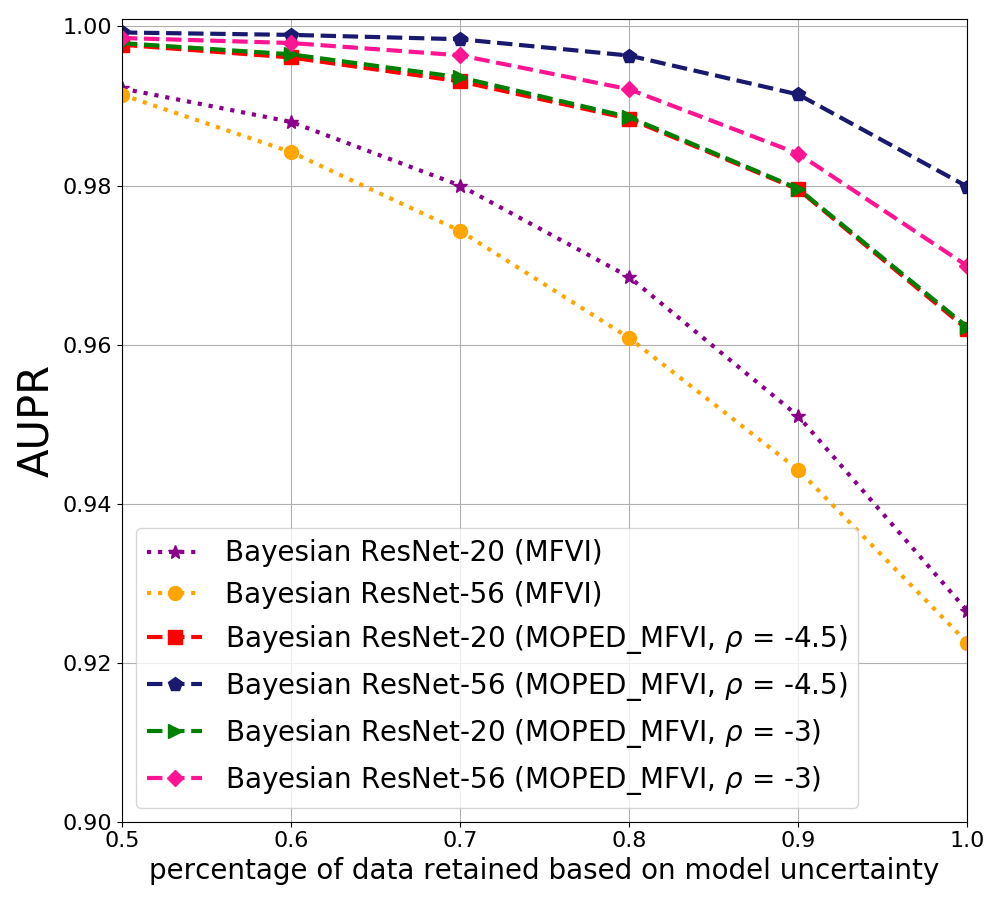}
		\caption{\small AUPR curves}
	\end{subfigure}
	\begin{subfigure}{0.295\textwidth}
		\centering
		\captionsetup{
			justification=centering}
		\centering
		\includegraphics[scale=0.225]{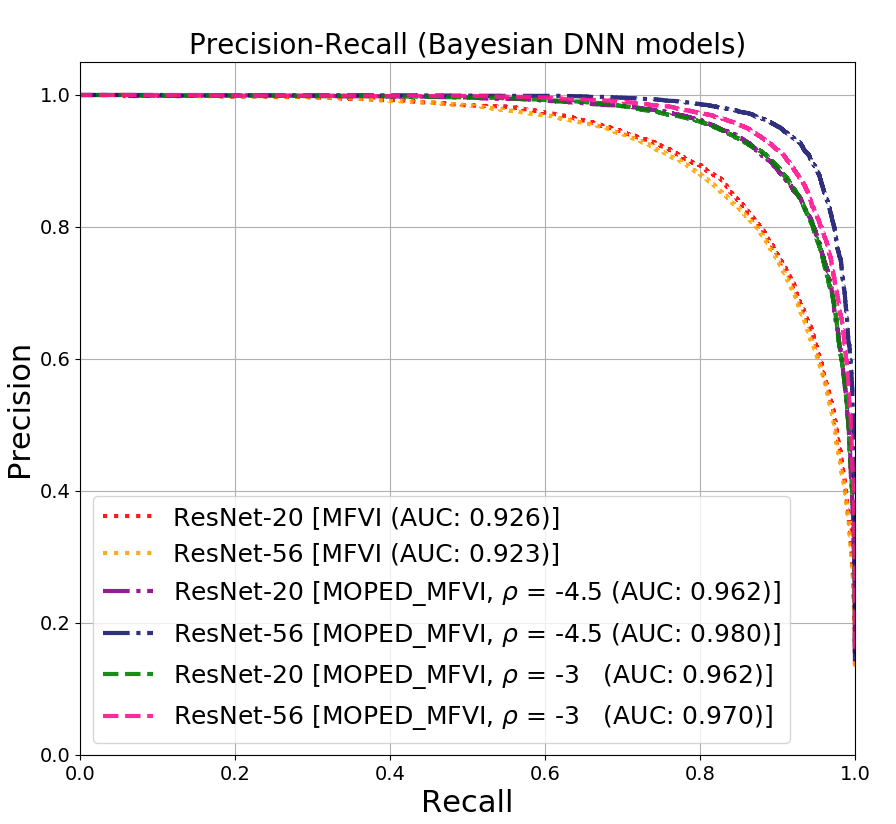}
		\caption{\small Precision-recall}
	\end{subfigure}
		\caption{\small Comparison of MOPED and MOPED\_MFVI for Bayesian ResNet-20 and ResNet-56 architectures. (a)~training convergence, (b)~AUPR as a function of retained data based on model uncertainty and (c)~precision-recall plots.
    }
	\label{fig:cifar-10-plots}
\end{figure*}

\begin{figure}[t]
	\small
	\begin{subfigure}{0.9\textwidth}
		\centering
		\captionsetup{
			justification=centering}
		\includegraphics[scale=0.23]{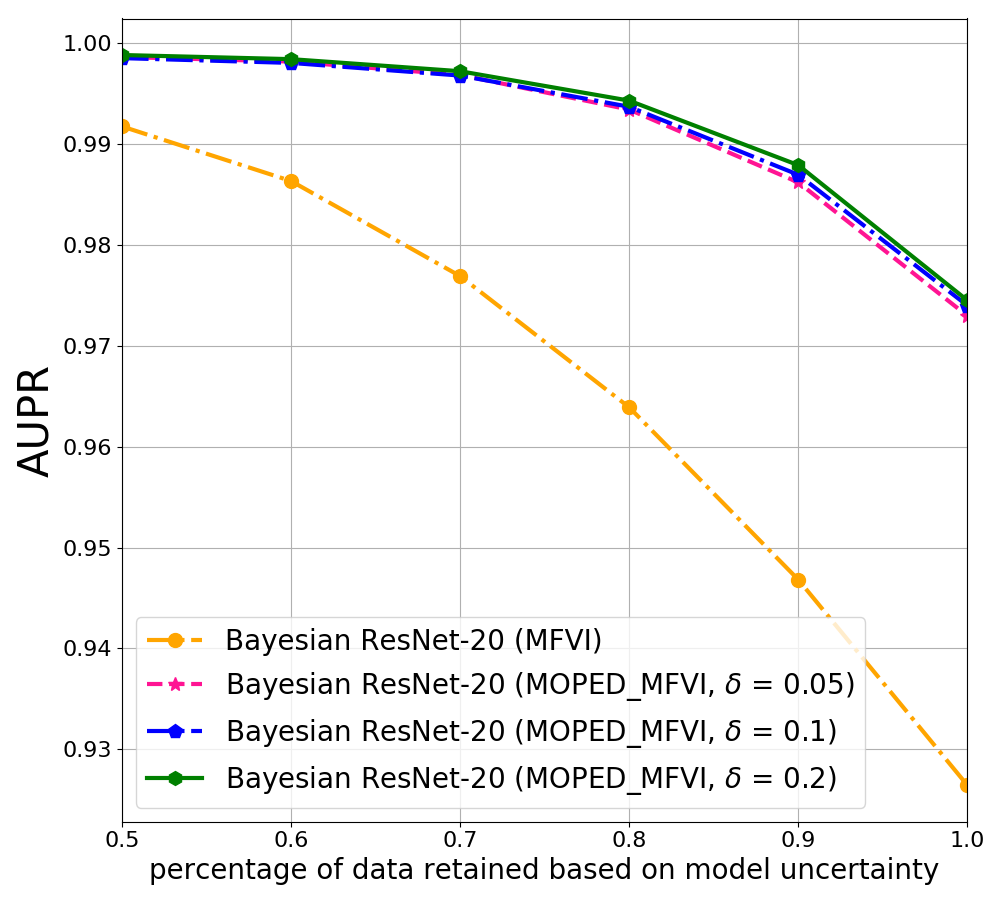}
		\caption{\small Bayesian ResNet-20 (CIFAR-10)}
	\end{subfigure}
	\begin{subfigure}{0.9\textwidth}
		\centering
		\captionsetup{
			justification=centering}
		\includegraphics[scale=0.24]{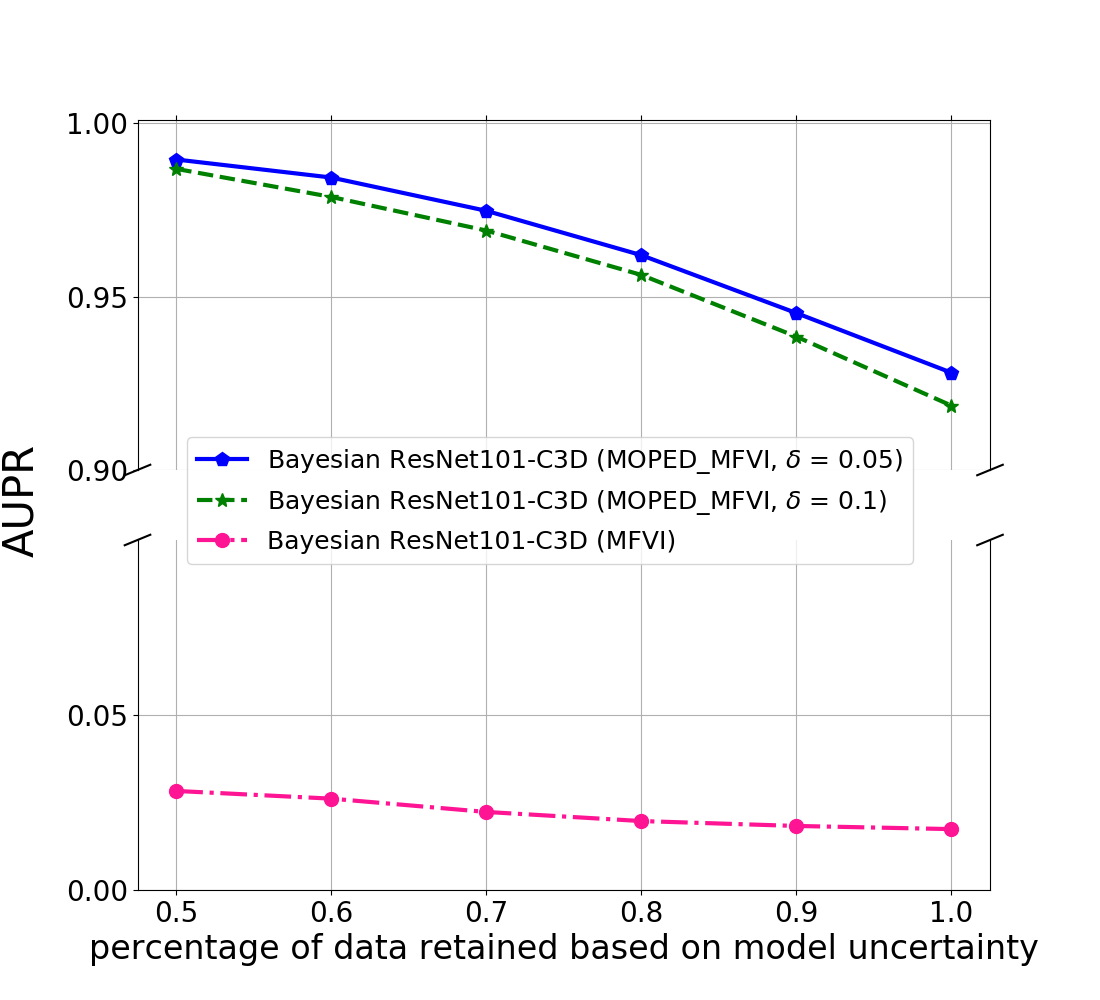}
		\caption{\small Bayesian ResNet-101 C3D (UCF-101)}
	\end{subfigure}
	\caption{\small Precision-recall AUC (AUPR) plots with different $\delta$ scale factors for initializing variance values in MOPED method.  }
	\label{fig:scaleplots}
\end{figure}
In the next section, we demonstrate the benefits of MOPED method for variational inference with extensive empirical experiments. We showcase the proposed MOPED method helps Bayesian DNN architectures to achieve better model performance along with reliable uncertainty estimates. 

\section{Related Work}
\label{sec:related_work}
Deterministic pretraining ~\cite{molchanov2017variational,sonderby2016ladder} has been used to improve model training for variational probabilistic models.~\citeauthor{molchanov2017variational} use a pretrained deterministic network for Sparse Variational Dropout method. 
~\citeauthor{sonderby2016ladder} use a warm-up method for variational-auto encoder by rescaling the KL-divergence term with a scalar term $\beta$, which is increased linearly from 0 to 1 during the first N epochs of training. Whereas, in our method we use the point-estimates from pretrained standard DNN of the same architecture to set the informed priors and the model is optimized with MFVI using full-scale KL-divergence term in ELBO.

Choosing weight priors in Bayesian neural networks 
is an active area of research. ~\citeauthor{atanov2018deep} propose implicit priors for variational inference in convolutional neural networks that exploit generative models. ~\citeauthor{nguyen2017variational} use prior with zero mean and unit variance, and  initialize the optimizer at the mean of the MLE model and a very small initial variance for small-scale MNIST experiments. ~\citeauthor{wu2018deterministic} modify ELBO with a deterministic approximation of reconstruction term and use Empirical Bayes procedure for selecting variance of prior in KL term (with zero prior mean). The authors also caution about inherent scaling of their method could potentially limit its practical use for networks with large hidden size. All of these works have been demonstrated only on small-scale models and simple datasets like MNIST. In our method, we retain the stochastic property of the expected log-likelihood term in ELBO, and specify both mean and variance of weight priors based on pretrained DNN with Empirical Bayes. Further, we demonstrate our method on large-scale Bayesian DNN models with complex datasets on real-world tasks.

\section{Experiments}

\label{sec:results}

We evaluate proposed method on real-world applications including image and audio classification, and video activity recognition. 
We consider multiple architectures with varying complexity to show the scalability of method in training deep Bayesian models. 
Our experiments include: (i) ResNet-101 C3D~\cite{hara3dcnns} for video activity classification on UCF-101\cite{soomro2012ucf101} dataset, (ii) VGGish\cite{hershey2017cnn} for audio classification on UrbanSound8K~\cite{salamon2014dataset} dataset, (iii) Modified version of VGG\cite{oatml2019bdlb} for diabetic retinopathy detection~\cite{diabetic_ret_challenge}, (iv) ResNet-20 and ResNet-56~\cite{he2016deep} for CIFAR-10~\cite{krizhevsky2009learning}, (v) LeNet architecture for MNIST~\cite{lecun1998gradient} digit classification, and (vi) Simple convolutional neural network (SCNN) consisting of two convolutional  layers followed by two dense layers for image classification on  Fashion-MNIST~\cite{xiao2017fashion} datasets.

We  implemented above Bayesian DNN models and trained them using Tensorflow and Tensorflow-Probability~\cite{dillon2017tensorflow} frameworks. The variational layers are modeled using Flipout~\cite{wen2018flipout}, an efficient method that decorrelates the gradients within a mini-batch by implicitly sampling pseudo-independent weight perturbations for each input. The MLE weights obtained from the pretrained DNN models are used in MOPED method to set the priors and initialize the variational parameters in approximate posteriors (Equation~\ref{eq:mean_sel} and~\ref{eq:scale_prior}), as described in Section\ref{sec:MOPED}. 

During inference phase, predictive distributions are obtained by performing multiple stochastic forward passes over the network while sampling from posterior distribution of the weights (40 Monte Carlo samples in our experiments). 
We evaluate the model uncertainty and predictive uncertainty using Bayesian active learning by disagreement (BALD)~(Equation~\ref{eq:mutual information}) and predictive entropy (Equation~\ref{eq:pred_entropy}), respectively. 
Quantitative comparison of uncertainty estimates are made by calculating area under the curve of precision-recall (AUPR) values by retaining different percentages (0.5 to 1.0) of most certain test samples (i.e. ignoring most uncertain predictions based on uncertainty estimates).

\begin{table*}
	\begin{center}
		\begin{tabular}{ M{3.5cm}  M{1.5cm} M{1.5cm} M {0.1cm} M{1.5cm}  M{1.5cm} M {0.1cm} M{1.5cm} M{1.5cm}}	\\\hlineB{2.5}
			& \multicolumn{2}{c}{50\% data retrained} & &  \multicolumn{2}{c}{75\% data retrained}  & & \multicolumn{2}{c}{100\% data retrained} \\ \cline{2-3}\cline{5-6}\cline{8-9}
			Method & AUC   &     Accuracy    &  &      AUC   &   Accuracy & & AUC  &  Accuracy \\\hline
			MC Dropout & 0.878     &    0.913      && 0.852    &    0.871 && 0.821  &    0.845 \\
			Mean-field VI &  0.866    &     0.881 && 0.84      &    0.850 &  &0.821   &   0.843\\
			Deep Ensembles &   0.872 &   0.899 &&  0.849 &   0.861 && 0.818   &   0.846\\
			Deterministic &  0.849 &   0.861 &&  0.823 &   0.849 &&  0.82 &  0.842\\
			Ensemble MC Dropout & 0.881 &   0.924 &&  0.854 &   0.881 &&  0.825   &  0.853\\
			\textbf {MOPED Mean-field VI} & \textbf{0.912} &  \textbf{0.937} && \textbf{0.885} &  \textbf{0.914} && \textbf{0.883}   &  \textbf{0.857} \\\hline
			Random Referral & 0. 818 &  0.848 &&       0.820 &  0.843 &&  0.820  &  0.842 \\\hlineB{2.5}
		\end{tabular}
	\end{center}
	\caption{\small Comparison of Area under the receiver-operating characteristic curve (AUC) and classification accuracy as a function of retained data from the BDL benchmark suite. The proposed method demonstrates an improvement of superior performance compared to all the baseline models.}
	\label{tab:bdl_benchmark}
\end{table*}

In Table~\ref{tab:Accuracy}, classification accuracies for architectures with various model complexity are presented. 
Bayesian DNNs with priors initialized with MOPED method achieves similar or better predictive accuracies as compared to equivalent DNN models. 
Bayesian DNNs with random initialization of Gaussian priors has difficulty in converging to optimal solution for larger models (ResNet-101 C3D and VGGish). 
It is evident from these results that MOPED method guarantees the training convergence even for the complex models. 

In Figure~\ref{fig:cifar-10-plots}, comparison of mean field variational inference with MOPED method (MOPED\_MFVI) and mean field variational inference with random initialization of priors (MFVI) is shown for Bayesian ResNet-20 and ResNet-56 architectures trained on CIFAR-10 dataset.  The AUPR plots capture the precision-recall AUC values as a function of retained data based on the model uncertainty estimates.   
Figure~\ref{fig:cifar-10-plots}~(b)~\&~(c) show that MOPED\_MFVI provides better performance than MFVI. AUPR increases as most uncertain predictions are ignored based on the model uncertainty, indicating reliable uncertainty estimates.
We show the results for different selection of $\rho$ values (as shown in Equation~\ref{eq:mean_sel}). 

In Figure~\ref{fig:scaleplots}, we show AUPR plots for CIFAR-10 and UCF-101with different $\delta$ values as mentioned in Equation~\ref{eq:scale_prior}. 

\begin{figure*}[t]
	\small
	\begin{subfigure}{0.48\textwidth}
		\centering
		\captionsetup{
			justification=centering}
		\includegraphics[scale=0.5]{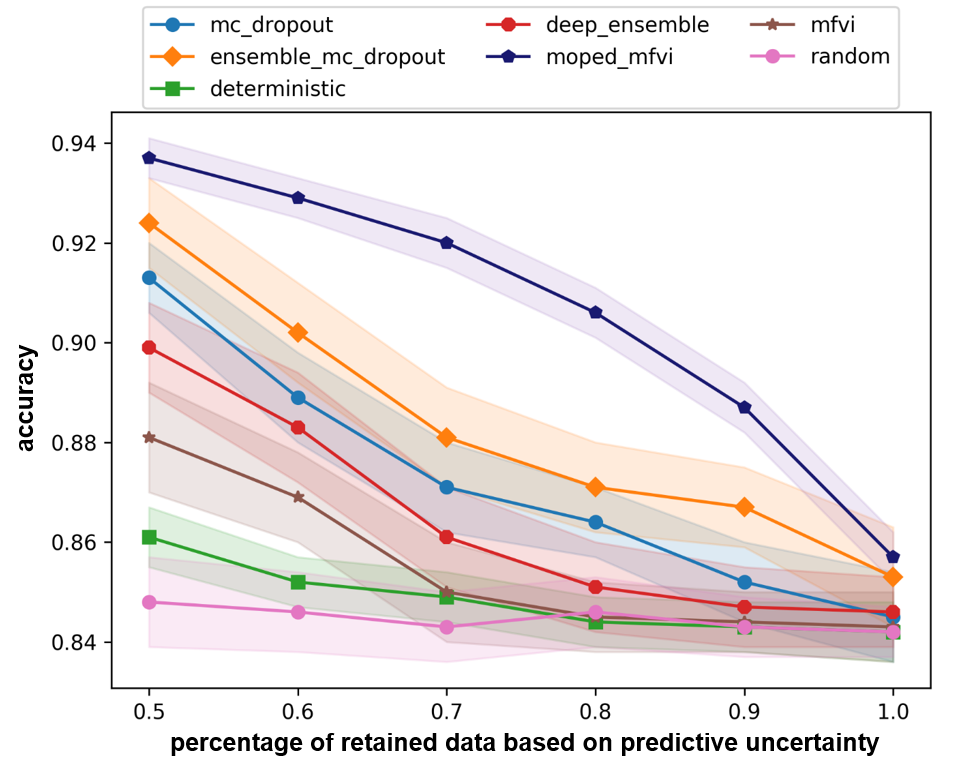}
		\caption{\small Binary Accuracy}
	\end{subfigure}
	\begin{subfigure}{0.48\textwidth}
		\centering
		\captionsetup{
			justification=centering}
		\includegraphics[scale=0.5]{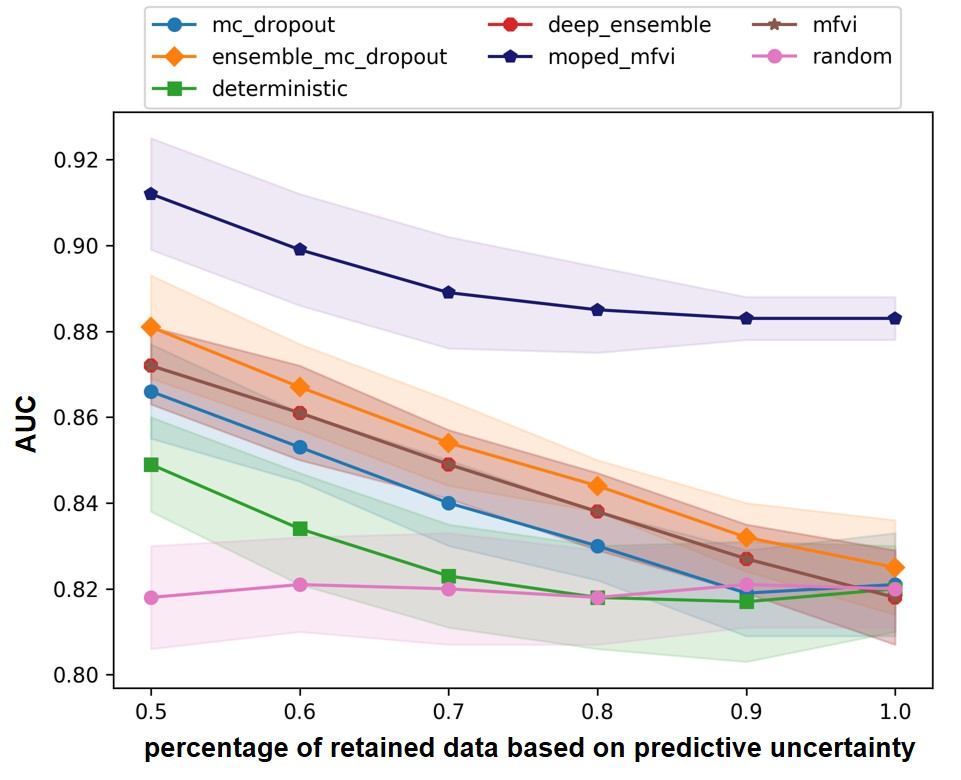}
		\caption{\small AUC-ROC}
	\end{subfigure}
	\caption{\small  Benchmarking MOPED\_MFVI with state-of-art Bayesian deep learning techniques on diabetic retinopathy diagnosis task using BDL-benchmarks. Accuracy and area under the receiver-operating characteristic curve (AUC-ROC) plots for varied percentage of retained data based on predictive uncertainty. MOPED\_MFVI performs better than the other baselines from BDL-benchmarks suite.  Shading shows the standard error. }
	\label{fig:diabetic_retinopathy}
\end{figure*}

\subsection {Benchmarking uncertainty estimates}
Bayesian Deep Learning (BDL) benchmarks~\cite{oatml2019bdlb} is an open-source framework for evaluating deep probabilistic machine learning models and their application to real-world problems. BDL-benchmarks assess both the scalability and effectiveness of different techniques for uncertainty estimation.  The proposed MOPED\_MFVI method is compared with state-of-the-art baseline methods available in BDL-benchmarks suite  on diabetic retinopathy detection task~\cite{diabetic_ret_challenge}. The evaluation methodology assesses the techniques by their diagnostic accuracy and area under receiver-operating-characteristic (AUC-ROC) curve, as a function of percentage of retained data based on predictive uncertainty estimates. It is expected that the models with well-calibrated uncertainty improve their performance (detection accuracy and AUC-ROC) as most certain data is retrained.

We have followed the evaluation methodology presented in the BDL-benchmarks to compare the accuracy and uncertainty estimates obtained from our method.  We used the same model architecture (VGG) and hyper-parameters as used by other baselines for evaluating MOPED\_MFVI. The results for the BDL baseline methods are obtained from~\cite{oatml2019bdlb}. In Table~\ref{tab:bdl_benchmark} and Figure~\ref{fig:diabetic_retinopathy}, quantitative evaluation of AUC and accuracy values for BDL baseline methods and MOPED\_MFVI are presented. The proposed MOPED\_MFVI method outperforms other state-of-the-art BDL techniques. 

\begin{figure}
	\begin{center}
		\includegraphics[scale=0.20]{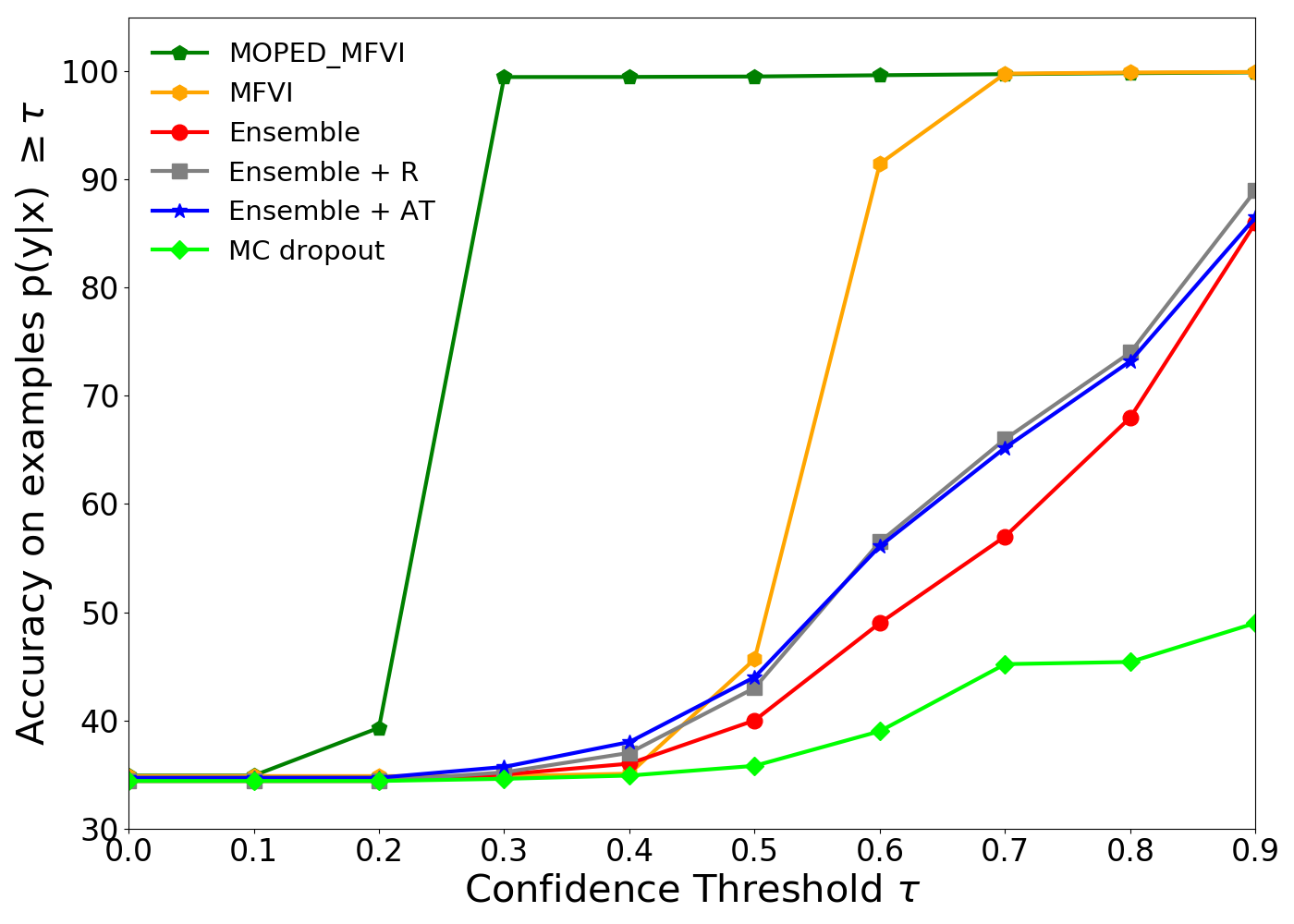}
	\end{center}
	\caption{\small Accuracy vs Confidence curves: Networks trained on MNIST and tested on both MNIST and the NotMNIST (out-of-distribution) test sets. }
	\label{fig:comparison_with_other_methods}
\end{figure}

\begin{figure*}
	\small
	\begin{subfigure}{0.24\textwidth}
		\centering
		\captionsetup{
			justification=centering}
		\includegraphics[scale=0.3]{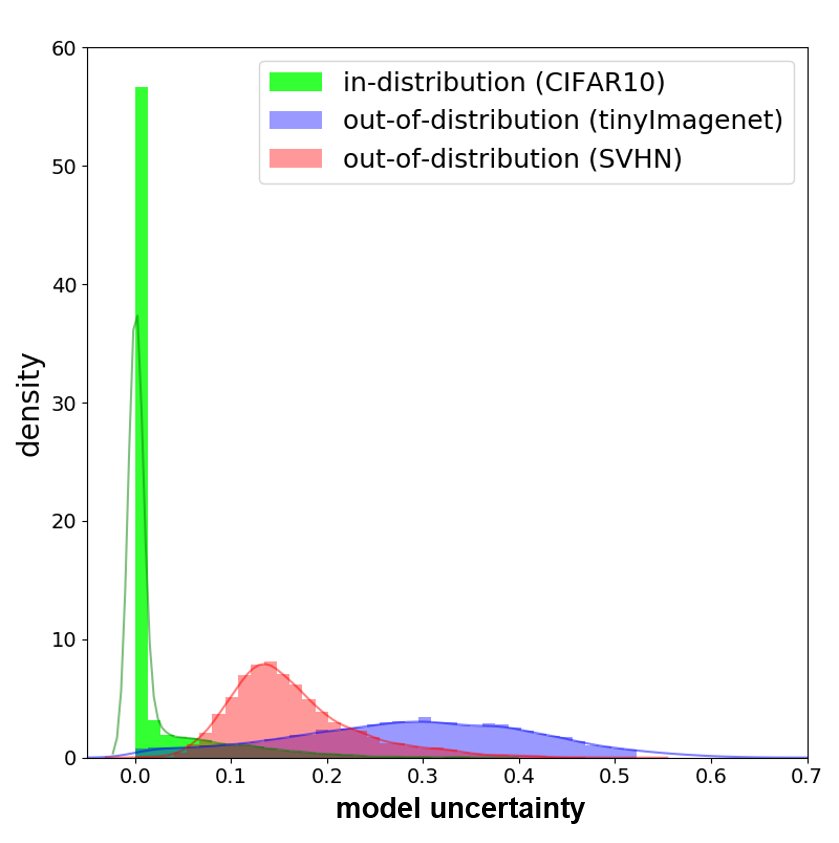}
		\caption{\small Model uncertainty ~~~~~~~~~(Bayesian ResNet-56) }
	\end{subfigure}
	\begin{subfigure}{0.24\textwidth}
		\centering
		\captionsetup{
			justification=centering}
		\includegraphics[scale=0.3]{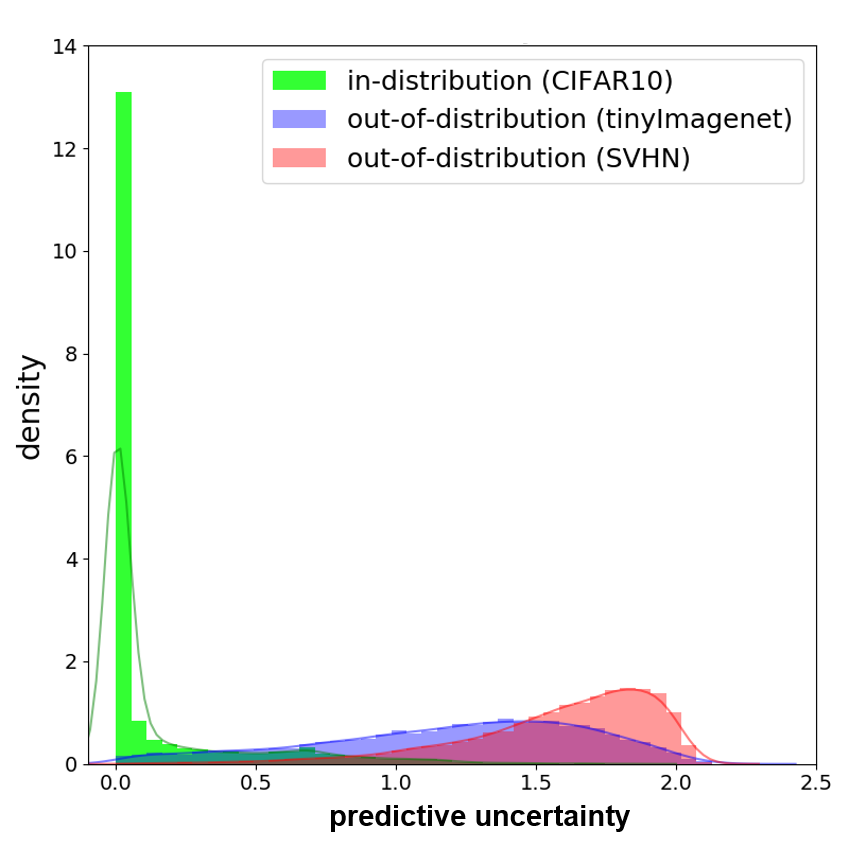}
		\caption{\small Predictive uncertainty (Bayesian ResNet-56) }
	\end{subfigure}
	\begin{subfigure}{0.24\textwidth}
		\centering
		\captionsetup{
			justification=centering}
		\includegraphics[scale=0.3]{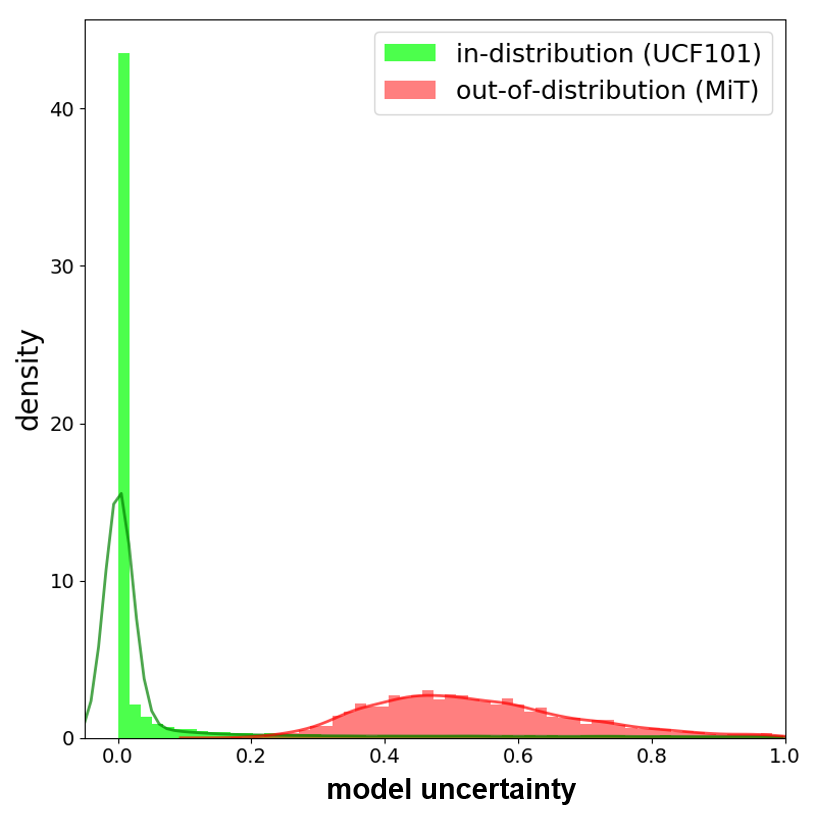}
		\caption{\small \small Model uncertainty ~~~~~~~~~(Bayesian ResNet-101 C3D)}
	\end{subfigure}
	\begin{subfigure}{0.24\textwidth}
		\centering
		\captionsetup{
			justification=centering}
		\includegraphics[scale=0.3]{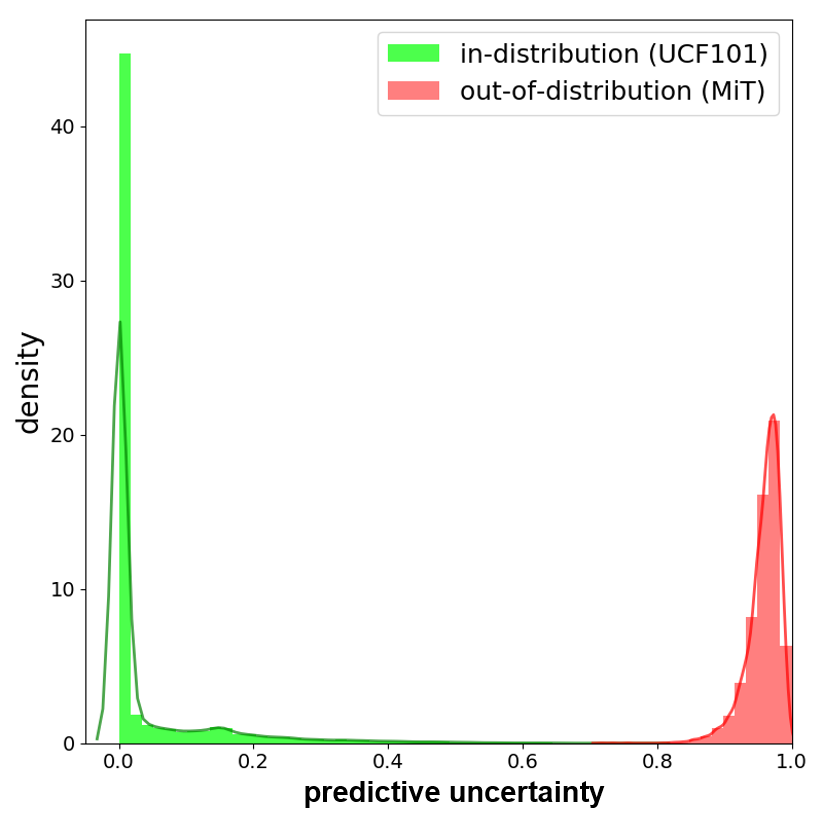}
		\caption{\small Predictive uncertainty (Bayesian ResNet-101 C3D)}
	\end{subfigure}
	\caption{\small Density histograms obtained from in- and out-of-distribution samples. Bayesian DNN model uncertainty estimates indicate higher uncertainty for out-of-distribution samples as compared to the in-distribution samples. }
	\label{fig:ood}
\end{figure*}

\subsection{Robustness to out-of-distribution data}
We evaluate the uncertainty estimates obtained from MOPED\_MFVI to detect out-of-distribution data. Out-of-distribution samples are data points which fall far off from the training data distribution.  
We evaluate two sets of out-of-distribution detection experiments. In the first set, we use CIFAR-10 as the in-distribution samples trained using ResNet-56 Bayesian DNN model. TinyImageNet~\cite{russakovsky2015imagenet}  and SVHN~\cite{goodfellow2013multi} datasets are used as out-of-distribution samples which were not seen during the training phase. The density histograms (area under the histogram is normalized to one) for uncertainty estimates obtained from the Bayesian DNN models are plotted in Figure~\ref{fig:ood}.   The density histograms in Figure~\ref{fig:ood}~(a) \& (b) indicate higher uncertainty estimates for the out-of-distribution samples and lower uncertainty values for the in-distribution samples. 
A similar trend is observed in the second set using UCF-101~\cite{soomro2012ucf101} and Moments-in-Time (MiT)~\cite{Monfort2018} video activity recognition datasets as the in- and out-of-distribution data, respectively. These results confirm the uncertainty estimates obtained from proposed  method are reliable and can identify out-of-distribution data.

In order to evaluate robustness of our method (MOPED\_MFVI), we compare state-of-the-art probabilistic deep learning methods for prediction accuracy as a function of model confidence. Following the experiments in~\cite{lakshminarayanan2017simple}, we trained our model on MNIST training set and tested it on a mix of examples from MNIST and NotMNIST 
(out-of-distribution) test set. The accuracy as a function of confidence plots should increase monotonically, as higher accuracy is expected for more confident results. A robust model should provide low confidence for out-of-distribution samples while providing high confidence for correct prediction from in-distribution samples. The proposed variational inference method with MOPED priors provides more robust results as compared to the MC Dropout~\cite{gal2016dropout} and deep model ensembles~\cite{lakshminarayanan2017simple} approaches (shown in Figure~\ref{fig:comparison_with_other_methods}).

\section{Conclusions}
We proposed MOPED method that specifies informed weight priors in Bayesian deep neural networks with Empirical Bayes approach. We demonstrated with thorough empirical experiments that MOPED enables scalable variational inference for Bayesian DNNs.
We demonstrated the proposed method outperforms state-of-the-art Bayesian deep learning techniques using BDL-benchmarks framework.
We also showed the uncertainty estimates obtained from the proposed method are reliable to identify out-of-distribution data. The results support proposed approach provides better model performance and reliable uncertainty estimates on real-world tasks with large scale complex models.

\bibliography{AAAI-KrishnanR.6158}
\bibliographystyle{aaai}
\end{document}